\documentclass{article} 
\usepackage{nips12submit_e,times}
\usepackage{graphicx}  
\usepackage{subfig}
\usepackage[linesnumbered,ruled]{algorithm2e}
\usepackage{amssymb,amsmath}
\usepackage{amsfonts}
\usepackage[multiple]{footmisc}
\nipsfinalcopy

\title{Communication-Efficient Parallel Belief Propagation for Latent Dirichlet Allocation}

\author{
Jian-Feng Yan\thanks{This work is supported by NSFC (Grant No. 61003154 and 61003259), a GRF grant from RGC UGC Hong Kong (GRF Project No.9041574), a grant from City University of Hong Kong (Project No. 7008026) and a grant from Baidu.} \\
Department of CS\\
Soochow University\\
Suzhou, China 215006 \\
\And
Zhi-Qiang Liu \\
School of Creative Media City \\University of Hong Kong\\
Tat Chee Ave 83 \\
\AND
Yang Gao\\
Department of CS\\
Soochow University\\
Suzhou, China 215006 \\
\And
Jia Zeng\\
Department of CS\\
Soochow University\\
Suzhou, China 215006 \\
\texttt{j.zeng@ieee.org} \\
}

\begin{document}

\maketitle

\begin{abstract}


This paper presents a novel communication-efficient parallel belief propagation (CE-PBP) algorithm for training latent Dirichlet allocation (LDA).
Based on the synchronous belief propagation (BP) algorithm, we first develop a parallel belief propagation (PBP) algorithm on the parallel architecture.
Because the extensive communication delay often causes a low efficiency of parallel topic modeling, we further use Zipf's law to reduce the total communication cost in PBP.
Extensive experiments on different data sets demonstrate that CE-PBP achieves a higher topic modeling accuracy
and reduces more than $80\%$ communication cost than the state-of-the-art parallel Gibbs sampling (PGS) algorithm.

\end{abstract}

\section{Introduction}
Topic modeling for massive data sets has attracted intensive research interests recently,
because large-scale data sets such as collections of images and documents are becoming increasingly common~\cite{Newman:09,Canini:09,Zhai:11,Hoffman:10}.
Online and parallel topic modeling algorithms have been two major strategies for massive data sets.
The former processes the massive data stream by mini-batches,
and discards the processed mini-batch after one look~\cite{Canini:09,Hoffman:10}.
The latter uses the parallel architecture to speed up the topic modeling by multi-core/processor and more memory resources~\cite{Newman:09,Zhai:11}.
Although online topic modeling algorithms use less computational resources,
their topic modeling accuracy depends on several heuristic parameters including the mini-batch size~\cite{Canini:09,Hoffman:10},
and is often comparable or less than batch learning algorithms.
In practice,
online algorithms are often $2\sim5$ times faster than batch algorithms~\cite{Hoffman:10},
while parallel algorithms can get $700$ times faster under $1024$ processors~\cite{Newman:09}.
Because the parallel architecture becomes cheaper and widely-used,
the parallel topic modeling algorithms are becoming an ideal choice to speed up topic modeling.
However,
parallel topic modeling is not a trivial task,
because its efficiency depends highly on extensive communication/synchrononization delays across distributed processors.
Indeed,
the communication cost determines the scalability of the parallel topic modeling algorithms.

In this paper, we propose a novel communication-efficient parallel belief propagation (CE-PBP) algorithm
for training latent Dirichlet allocation (LDA)~\cite{Blei:03}, one of the simplest topic models.
First, we extend the synchronous BP algorithm~\cite{Zeng:11} to PBP on the parallel architecture for training LDA.
We show that PBP can yield exactly the same results as the synchronous BP.
Second,
to reduce extensive communication/synchrononization delays,
we use Zipf's law~\cite{newman2005power} to determine the communication rate of synchronizing global parameters in PBP.
Using the different communication rates, we show that the total communication cost can be significantly reduced.
Extensive experiments confirm that CE-PBP reduces around $85\%$ communication time,
and achieves a much higher topic modeling accuracy than the state-of-the-art parallel Gibbs sampling algorithm (PGS)~\cite{Newman:09,wang2009plda}.

\section{Parallel Belief Propagation (PBP)} \label{s2}

LDA allocates a set of semantic topic labels,
$\mathbf{z} = \{z^k_{w,d}\}$,
to explain non-zero elements
in the document-word co-occurrence matrix $\mathbf{x}_{W \times D} = \{x_{w,d}\}$,
where $1 \le w \le W$ denotes the word index in the vocabulary,
$1 \le d \le D$ denotes the document index in the corpus,
and $1 \le k \le K$ denotes the topic index.
Usually,
the number of topics $K$ is provided by users.
The topic label satisfies $z^k_{w,d} = \{0,1\}, \sum_{k=1}^K z^k_{w,d} = 1$.
After inferring the topic labeling configuration over the document-word matrix,
LDA estimates two matrices of multinomial parameters:
topic distributions over the fixed vocabulary $\boldsymbol{\phi}_{W \times K} = \{\phi_{\cdot,k}\}$,
where $\theta_{\cdot, d}$ is a $K$-tuple vector and $\phi_{\cdot,k}$ is a $W$-tuple vector,
satisfying $\sum_k \theta_{k,d} = 1$ and $\sum_w \phi_{w,k} = 1$.
From a document-specific proportion $\theta_{\cdot,d}$,
LDA generates a topic label $z^k_{\cdot,d}=1$,
which further combines $\phi_{\cdot,k}$ to generate a word index $w$,
forming the total number of observed word counts $x_{w,d}$.
Both multinomial vectors $\theta_{\cdot,d}$ and $\phi_{\cdot,k}$ are generated by two Dirichlet distributions with hyperparameters $\alpha$ and $\beta$.
For simplicity,
we consider the smoothed LDA with fixed symmetric hyperparameters provided by users~\cite{Griffiths:04}.

After integrating out the multinomial parameters $\{\phi,\theta\}$,
LDA becomes the collapsed LDA in the collapsed hidden variable space $\{\mathbf{z, \alpha, \beta}\}$.
The collapsed Gibb sampling (GS)~\cite{Griffiths:04} is a Markov Chain Monte Carlo (MCMC) sampling technique to infer the marginal distribution or {\em message},
$\mu_{w,d,i}(k) = p(z^k_{w,d,i}=1)$,
where $1 \le i \le x_{w,d}$ is the word token index.
The message update equation is
\begin{align} \label{GS}
\mu_{w,d,i}(k) \propto (\mathbf{z}^k_{\cdot,d,-i} + \alpha) \times
\frac{\mathbf{z}^k_{w,\cdot,-i} + \beta}{\sum_w (\mathbf{z}^k_{w,\cdot,-i} + \beta)},
\end{align}
where $\mathbf{z}^k_{\cdot,d,-i} = \sum_{w} z^k_{w,d,-i}$,
$\mathbf{z}^k_{w,\cdot,-i} = \sum_{d} z^k_{w,d,-i}$,
and the notation $-i$ denotes excluding the current topic label $z^k_{w,d,i}$.
Then,
GS randomly samples a topic label $z^k_{w,d,i}=1$ from the message,
and immediately estimates messages of other word tokens.

Unlike GS,
BP~\cite{Zeng:11} infers messages,
$\mu_{w,d}(k) = p(z^k_{w,d}=1)$,
without sampling in order to keep all uncertainties of messages.
The message update equation is
\begin{align} \label{BP}
\mu_{w,d}(k) \propto [\boldsymbol{\mu}_{-w,d}(k) + \alpha] \times
\frac{\boldsymbol{\mu}_{w,-d}(k) + \beta}{\sum_w[\boldsymbol{\mu}_{w,-d}(k) + \beta]},
\end{align}
where $\boldsymbol{\mu}_{-w,d}(k) = \sum_{-w} x_{-w,d}\mu_{-w,d}(k)$ and
$\boldsymbol{\mu}_{w,-d}(k) = \sum_{-d} x_{w,-d}\mu_{w,-d}(k)$.
The notation $-w$ and $-d$ denote all word indices except $w$ and all document indices except $d$.
Eq.~\eqref{BP} differs from Eq.~\eqref{GS} in two aspects.
First,
BP infers messages based on word indices rather than word tokens.
Second,
BP updates and passes complete messages without sampling.
In this sense,
BP can be viewed as a {\em soft} version of GS.
Obviously,
such differences give Eq.~\eqref{BP} two advantages over Eq.~\eqref{GS}.
First,
it keeps all uncertainties of messages for higher topic modeling accuracy.
Second,
it scans the number of non-zero elements ($NNZ$) for message passing,
which is significantly less than the total number of word tokens $\sum_{w,d} x_{w,d}$ in $\mathbf{x}$.
So,
BP is often faster than GS by scanning a significantly less number of elements ($NNZ \ll \sum_{w,d}x_{w,d}$) at each training iteration~\cite{Zeng:11}.

Based on the parallel architecture,
we propose the parallel belief propagation (PBP) algorithm to speed up the synchronous BP.
First,
we define two matrices,
\begin{align}
\label{theta}
\hat{\boldsymbol{\theta}}_{k,d} = \sum_{w} x_{w,d}\mu_{w,d}(k), \\
\label{phi}
\hat{\boldsymbol{\phi}}_{w,k} = \sum_{d} x_{w,d}\mu_{w,d}(k),
\end{align}
so that we can re-write~\eqref{BP} as
\begin{align} \label{PBP}
\mu_{w,d}(k) \propto [\hat{\boldsymbol{\theta}}^{-w}_{k,d} + \alpha] \times
\frac{\hat{\boldsymbol{\phi}}^{-d}_{w,k} + \beta}{\sum_w\hat{\boldsymbol{\phi}}^{-d}_{w,k} + W\beta},
\end{align}
where $-w$ and $-d$ denote excluding $x_{w,d}\mu_{w,d}(k)$ from the matrices~\eqref{theta} and~\eqref{phi}.
At each training iteration $t, 1 \le t \le T$,
the synchronous BP updates the message~\eqref{PBP} using~\eqref{theta} and~\eqref{phi}
at $t-1$ iteration for non-zero elements in $\mathbf{x}$.
The updated messages are then used to estimate two matrices~\eqref{theta} and~\eqref{phi} at $t$ iteration.
After $T$ iterations,
the synchronous BP stops and normalizes $\sum_k \hat{\boldsymbol{\theta}}_{k,d} = 1$ and $\sum_w \hat{\boldsymbol{\phi}}_{w,k} = 1$ to
obtain the multinomial parameters $\boldsymbol{\theta}_{k,d}$ and $\boldsymbol{\phi}_{w,k}$.

PBP distributes $D$ documents into $1 \le m \le M$ processors.
Thus,
the matrix $\hat{\boldsymbol{\theta}}_{K \times D}$ can be also distributed into $M$ processors as $\hat{\boldsymbol{\theta}}_{k,d,m}$,
but the matrix $\hat{\boldsymbol{\phi}}_{W \times K}$ is shared by $M$ processors.
At each training iteration $t$,
each local processor $m$ sweeps the local data $\mathbf{x}_{w,d,m}$ using Eqs.~\eqref{theta} to~\eqref{PBP}.
The updated local $\hat{\boldsymbol{\theta}}_{k,d,m}$ is independent,
but the updated local $\hat{\boldsymbol{\phi}}_{w,k,m}$ should influence each other across $M$ processors.
So,
we need to communicate each local $\hat{\boldsymbol{\phi}}_{w,k,m}$ in order to synchronize the global matrix $\hat{\boldsymbol{\phi}}_{w,k}$,
\begin{align} \label{syn1}
\hat{\boldsymbol{\phi}}_{w,k} \leftarrow \hat{\boldsymbol{\phi}}_{w,k} + \sum_{m=1}^M(\hat{\boldsymbol{\phi}}_{w,k,m} - \hat{\boldsymbol{\phi}}_{w,k}),
\end{align}
After synchronization,
we copy the global $\hat{\boldsymbol{\phi}}_{w,k}$ to local $\hat{\boldsymbol{\phi}}_{w,k,m}$ for the next training iteration,
\begin{align} \label{syn2}
\hat{\boldsymbol{\phi}}_{w,k,m} \leftarrow \hat{\boldsymbol{\phi}}_{w,k}.
\end{align}
Because PBP follows the synchronous schedule,
it produces exactly the same results of the synchronous BP~\cite{Zeng:11}.
Notice that the parallel Gibbs sampling algorithm (PGS)~\cite{Newman:09,wang2009plda} is an approximate solution to GS in~\eqref{GS},
because GS uses an asynchronous schedule for message passing~\cite{Zeng:11}.

According to Eqs.~\eqref{syn1} and~\eqref{syn2},
PBP needs to communicate and synchronize a total of $2 \times M \times \hat{\boldsymbol{\phi}}_{K \times W}$ matrices at each training iteration $t$.
Let us take WIKI data set in Table~\ref{corpus} as an example.
If $K=10$ and $M=32$,
we need to communicate $400$MBytes in the parallel architecture at each training iteration.
This communication cost is so high as to delay synchronization.
For a total of $T$ training iterations,
the total communication cost is calculated as
\begin{align} \label{communication}
\text{Total communication cost} = 2 \times M \times T \times \hat{\boldsymbol{\phi}}_{K \times W}.
\end{align}
Notice that the communication cost of PGS~\cite{Newman:09,wang2009plda} can be also calculated as~\eqref{communication},
but with the following major difference.
In a common $32$-bit desktop computer,
PBP uses the double type ($8$ byte) but PGS uses the integer type ($4$ byte) to store the matrix $\hat{\boldsymbol{\phi}}_{W \times K}$.
So,
PGS requires only half communication cost as PBP,
i.e.,
around $200$MBytes in the above WIKI example.
Because this communication cost is still a bottleneck,
to reduce~\eqref{communication},
PGS changes the communication rate by running~\eqref{syn1} and~\eqref{syn2} at every $T' > 1$
training iterations~\cite{Newman:09}, so that the total communication cost can be reduced to a fraction $1/T'$ of~\eqref{communication}.
However,
the low communication rate slows down the convergence
and degrades the overall topic modeling performance of PGS~\cite{Newman:09}.
As a result,
PGS suggests running~\eqref{syn1} and~\eqref{syn2} at every training iteration,
which causes a serious communication delay.
In CE-PBP,
we aim to reduce the total communication cost~\eqref{communication} using Zipf's law without degrading the overall topic modeling performance very much.

\section{Reduce Communication Costs by Zipf's Law} \label{s3}

\begin{figure}[tb]
  \centering
  \subfloat{\includegraphics[scale=0.4]{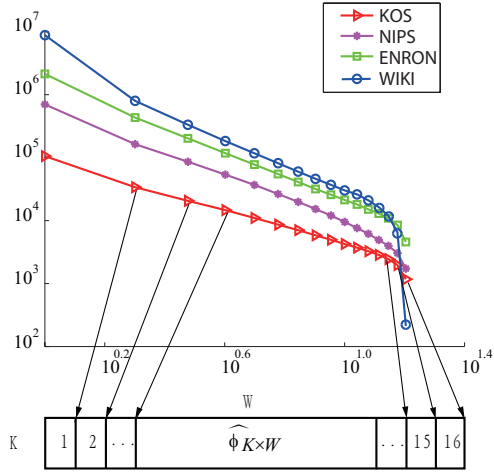}\label{zipf-2-3}}\qquad
  \caption{Zipf's law regulates different communication rates of different parts.
The top panel shows four Zipf's curves for the four data sets in Table~\ref{corpus}.
The bottom panel shows the uniform partition of the matrix $\hat{\boldsymbol{\phi}}_{W \times K}$ into $16$ parts according to Zipf's curves.
In case of $T=16$ training iterations,
the part with rank $1$ communicates $16$ times,
while the part with rank $16$ communicates only once when $H=1$.}
  \label{zipf}
\end{figure}

Zipf's law~\cite{zipf1949human} reveals that the word frequency rank $r$
has the following relationship with the word frequency $f$ in many natural language data sets ,
\begin{align} \label{zipflaw}
\log r = C - H \log f,
\end{align}
where $C$ and $H$ are positive constants.
Zipf's law indicates that the logarithm of the word rank in the frequency table is inversely proportional to its logarithm of frequency.
Generally,
the frequency of a word determines its contribution to message passing as shown in~\eqref{theta} and~\eqref{phi}.
So,
the higher word rank corresponds to the more contribution to topic modeling.

\IncMargin{3em}
\begin{algorithm}[t]\label{CE-PBP}
\caption{The CE-PBP Algorithm.}
\KwIn{$\mathbf{x}, T, K, M, N, \alpha, \beta $}
\KwOut{$\hat{\boldsymbol{\phi}}_{W \times K}, \hat{\boldsymbol\theta}_{K \times D}$}
Distribute $x_{w,d,m}$ to $M$ processors\;
Random initialization: global $\hat{\boldsymbol{\phi}}^0_{w,k}$ and local $\hat{\boldsymbol\theta}^0_{k,d,m}$\;
Copy global $\hat{\boldsymbol{\phi}}^0_{w,k}$ to local processor:
$\hat{\boldsymbol{\phi}}^0_{w,k,m} \leftarrow \hat{\boldsymbol{\phi}}^0_{w,k}$\;
\For{$t \leftarrow 1$ to $T$}
{
  \For{each processor $m$ in parallel} {
  \For{each part with rank $r$ in communication by Zipf's law} {
  Update local sub-matrices: $\hat{\boldsymbol{\phi}}^t_{\frac{W}{N},k,m} \leftarrow \hat{\boldsymbol{\phi}}^{t-1}_{\frac{W}{N},K}$\;
  }
  \For{$d \gets 1$ \KwTo $D$, $w \gets 1$ \KwTo $W$, $k \gets 1$ \KwTo $K$, $x_{w,d,m} \ne 0$}{
  $\mu^t_{w,d,m}(k) \propto [\hat{\boldsymbol{\theta}}^{-w}_{k,d,m} + \alpha] \times
   \frac{\hat{\boldsymbol{\phi}}^{-d}_{w,k,m} + \beta}{\sum_w\hat{\boldsymbol{\phi}}^{-d}_{w,k,m} + W\beta}$\;
  }
  $\hat{\boldsymbol{\theta}}^{t}_{k,d,m} = \sum_{w} x_{w,d,m}\mu^t_{w,d,m}(k)$ \;
  $\hat{\boldsymbol{\phi}}^{t}_{w,k,m} = \sum_{d} x_{w,d,m}\mu^t_{w,d,m}(k)$ \;
  }
  \tcp{Zipf's law based communication and synchronization}
  \For{each part with rank $r$ in communication by Zipf's law} {
  Update global sub-matrices:
  $\hat{\boldsymbol{\phi}}_{\frac{W}{N},K} \leftarrow \hat{\boldsymbol{\phi}}_{\frac{W}{N},K} +
  \sum_{m=1}^M(\hat{\boldsymbol{\phi}}_{\frac{W}{N},K,m} - \hat{\boldsymbol{\phi}}_{\frac{W}{N},K}),$\;
  }
}
\end{algorithm}
\DecMargin{3em}

To reduce the total communication cost~\eqref{communication},
we aim to reduce the size of matrix $\hat{\boldsymbol{\phi}}_{W \times K}$ at each training iteration.
First,
we uniformly partition the matrix $\hat{\boldsymbol{\phi}}_{W \times K}$ in terms of $W$ into $N$ parts,
where the first part with rank $r = 1$ contains the most frequent words in the vocabulary and so on.
If we sort the word frequency for different parts in a descending order,
we can plot the approximate Zipf's curves for the four data sets in Table~\ref{corpus} according to~\eqref{zipflaw} as shown in Fig.~\ref{zipf}.
As a result,
we obtain $N$ sub-matrices $\hat{\boldsymbol{\phi}}_{\frac{W}{N} \times K}$, satisfying $W/N \ll W$.
Here,
we use different communication rates for different parts,
\begin{align} \label{rate}
\text{Communication rate} = r^H,
\end{align}
where $r$ is the part rank in terms of word frequency and $H$ is the slope of Zipf' curve in Eq.~\eqref{zipf}.
When the slope $H$ is large,
the word frequency is small in part with large rank $r$,
which has a low communication rate to save time.
When $H=1$,
the part with rank $r$ communicates $\lfloor T/r \rfloor$ times in $T$ training iterations,
where $\lfloor \cdot \rfloor$ is the floor operation.
As a result,
the part with rank $r$ starts communicating if and only if the current iteration $t$ is multiples of $r$.
For example,
when $T=100$,
the part with rank $16$ will communicate when the current iteration $t \in \{16, 32, 48, 64, 80, 96\}$.
Based on Eq.~\eqref{rate},
the total communication cost~\eqref{communication} becomes
\begin{align} \label{reduced}
\text{Reduced communication cost} = 2 \times M \times \frac{\sum_{r=1}^N \lfloor T/r^H \rfloor}{N} \times \hat{\boldsymbol{\phi}}_{W \times K},
\end{align}
where $\lfloor \cdot \rfloor$ is the floor operation.
Because we use different communication rates $\lfloor T/r^H \rfloor$ for different parts with rank $r$,
Eq.~\eqref{reduced} is significantly smaller than~\eqref{communication}, i.e.,
$\frac{\sum_{r=1}^N  \lfloor T/r^H \rfloor}{N} \ll T$.
Let us take WIKI data set in Table~\ref{corpus} as an example.
If $K=10$ and $M=32$,
we need to communicate/synchronize $200$GBytes in the parallel architecture for $T=500$ iterations according to~\eqref{communication}.
In our strategy,
if $N = 100$,
we require only around $10$GBytes for communication/synchronization according to~\eqref{reduced},
which takes only $5\%$ communication cost in~\eqref{communication}.

The Zipf's law based communication rates are reasonable because the global matrix $\hat{\boldsymbol{\phi}}_{W \times K}$
is dominated by the high frequent words in~\eqref{phi}.
So,
the high communicate rate for the sub-matrix with top word frequencies ensures the accuracy of the global $\hat{\boldsymbol{\phi}}_{W \times K}$.
Compared with the uniform communication rate for the entire matrix $\hat{\boldsymbol{\phi}}_{W \times K}$ in PGS,
the different communication rates for different sub-matrices $\hat{\boldsymbol{\phi}}_{\frac{W}{N} \times K}$ are more effective.
The experiments on several data sets confirm that the proposed communication method degrades only $1\%$ topic modeling accuracy measured by the training perplexity,
but gains much higher parallel topic modeling efficiency.

The communication-efficient parallel belief propagation (CE-PBP) algorithm is summarized in Algorithm~\ref{CE-PBP}.
From Line $1$ to $3$,
we distribute the document-word matrix $\mathbf{x}_{W \times D}$ into $M$ processors,
and randomly initialize the global and local parameter matrices.
During each training iteration $t$,
we perform the parallel message passing independently in $M$ processors.
At the end of each training iteration,
we communicate and synchronize the global parameter matrix according to Zipf's law based communication rates.
Therefore,
CE-PBP reduces the total communication cost by different communication rates for different sub-matrices $\hat{\boldsymbol{\phi}}_{\frac{W}{N} \times K}$.

\section{Experiments}

We use four data sets\footnote{http://archive.ics.uci.edu/ml/machine-learningdatabases}\footnote{http://nlp.uned.es/social-tagging/wiki10+}:
KOS, NIPS, ENRON and WIKI in Table~\ref{corpus},
where $D$ denotes the number of documents,
$W$ denotes the size of vocabulary,
$NNZ$ denotes the number of non-zero elements in the document-word matrix.
Since KOS is a relatively smaller data set,
we use it for parameter tuning in CE-PBP.
The number of topics, $K=100$, is fixed in all experiments except for special statements.
The number of training iterations $T=500$.
We use the same hyperparameters $\alpha  = \beta = 0.01$ for CE-PBP and PGS.
Due to limited computational resources,
we use only $32$ processors to compare the performance of CE-PBP and PGS.
We find that the communication cost follows Eqs.~\eqref{communication} and~\eqref{reduced},
so that our results can be safely generalized to more processors in the parallel architecture.

\begin{table}[t]
\caption{Statistics of four document data sets.}
\label{corpus}
\begin{center}
\begin{tabular}{llll}
\multicolumn{1}{c}{\bf Date Sets}  &\multicolumn{1}{c}{\bf D} &\multicolumn{1}{c}{\bf W} &\multicolumn{1}{c}{\bf NNZ}
\\ \hline \\
KOS              &3430    &6906    &353160\\
NIPS             &1740    &13649   &933295\\
ENRON            &39861   &28102   &3710420\\
WIKI             &20758   &83470   &9272290\\
\end{tabular}
\end{center}
\end{table}

\subsection{Parameters for CE-PBP}

\begin{figure}[htb]
  \centering
  \subfloat{\includegraphics[scale=0.65]{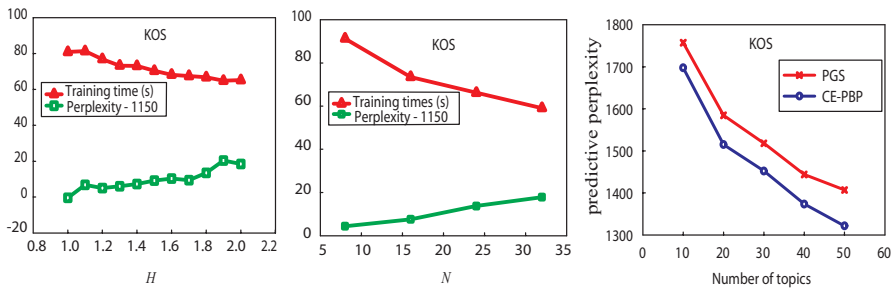}}\qquad
  \caption{Left (a): Predictive perplexity as a function of $H$.
  Middle (b): Predictive perplexity as a function of $N$.
  Right (c): Communication time of PGS, PBP and CE-PBP (seconds).}
  \label{hn}
\end{figure}

The parameter $H$ is the slope of the Zipf's curve in Fig.~\ref{zipf-2-3},
which determines the communication rate for part $r$.
Although Zipf's law applies to many natural language data sets,
some data sets do not fit Zipfian distribution perfectly,
which can be easily validated by $H$.
For example,
the parameter $H$ for KOS data set varies from $1$ to $1.6$.
Consequently,
we want to know if different $H$ will influence the performance of CE-PBP.
Fixing $N=16$,
we change $H$ from $1$ to $2$ with the step $0.1$ to investigate both training time and predictive perplexity,
where the predictive perplexity on test data set is calculated as in~\cite{Asuncion:08,Zeng:11}.
Usually,
the lower predictive perplexity often corresponds to the higher topic modeling accuracy.
Fig.~\ref{hn}a shows how training time and predictive perplexity change with the parameters $H$ on the KOS data set,
respectively.
Notice that we subtract $1150$ from the value of predictive perplexity to fit in the same figure.
Fig.~\ref{hn}a shows that when $H$ increases from $1$ to $2$,
the training time decreases slowly,
while predictive perplexity increases.
When $H=1$ in Eq.~\eqref{zipf},
CE-PBP achieves the highest accuracy with a slightly more training time.
So,
we empirically set $H=1$ in the rest of experiments.

On the other hand,
the number of parts $N$ for the global parameter matrix $\hat{\boldsymbol{\phi}}$ influences the topic modeling performance of CE-PBP.
The larger $N$ leads to the more communication cost reduction according to~\eqref{reduced}.
However,
the larger $N$ implies that in the fixed training iteration $T$ more sub-matrices communicate less frequently,
degrading the overall topic modeling performance.
Fixing $H=1$,
we change $N$ from $1$ to $32$ with the step size $8$ in the experiments.
Fig.~\ref{hn}b shows that the effect of parameter $N$.
While communication cost decreases with $N$ according to~\eqref{reduced},
the predictive perplexity increases steadily with $N$ because the part with higher rank $r$ communicates less frequently.
We empirically set $N=16$ to achieve a relatively balanced performance.
In this case,
the communication cost of CE-PBP is around $20\%$ of PBP according to~\eqref{communication} and~\eqref{reduced}.

Under the parameters $H=1$ and $N=16$,
we compare the predictive perplexity between CE-PBP and PGS in Fig.~\ref{hn}c.
When the number of topics $K\in\{10,20,30,40,50\}$,
CE-PBP consistently achieves much lower predictive perplexity values than PGS.
Such results are consistent with previous results on comparing BP and GS~\cite{Zeng:11}.
As a summary,
CE-PBP has a higher topic modeling accuracy than PGS in terms of perplexity.

\begin{figure}[t]
  \centering
  \includegraphics[scale=0.5]{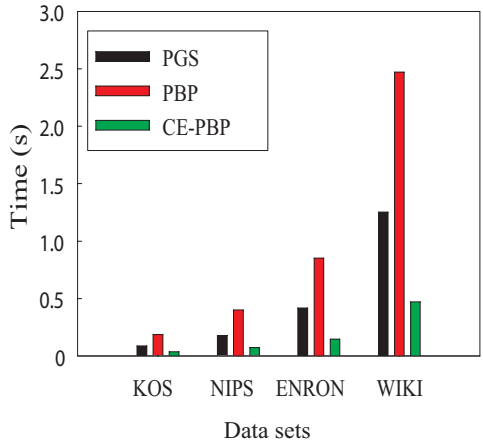}
  \caption{Communication time of PGS, PBP and CE-PBP (seconds).}
  \label{ccost}
\end{figure}

Fig.~\ref{ccost} compares the overall communication cost of PGS, PBP and CE-PBP on four data sets in Table~\ref{corpus}.
On the four data sets,
we find that the communication time of PBP is around twice that of PGS.
This result follows our analysis that PBP uses double type while PGS uses integer type to store the global parameter matrix.
The actual communication cost of CE-PBP is about $17\%$ of PBP,
slightly shorter than the expected value $20\%$ according to~\eqref{reduced}.
Such an improvement can be partly attributed to less input/output conflicts during updating the global parameter matrix.
Since the access time of the global parameter is remarkably reduced,
there are less access conflicts among all processors.

\begin{figure}[t]
  \centering
  \subfloat{\includegraphics[scale=0.6]{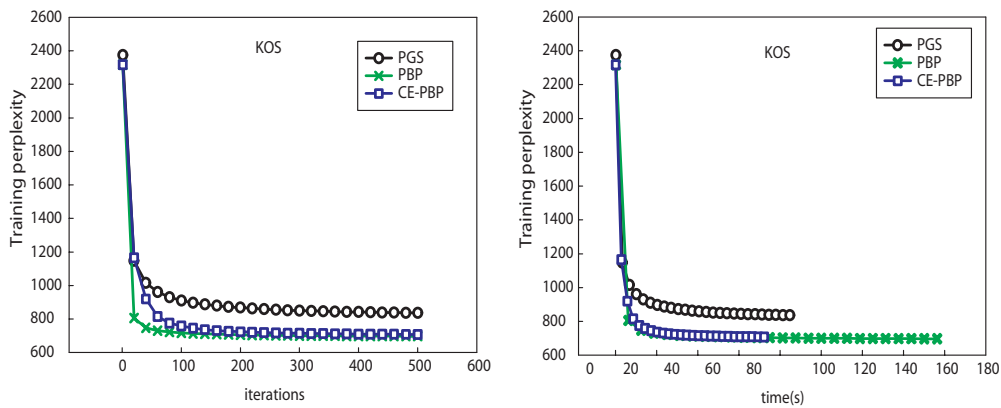}}\qquad
  \caption{(a) Left: Perplexity as a function of iteration on KOS.
  (b) Right: Perplexity as a function of training time on KOS.}
  \label{convergence}
\end{figure}

Fig.~\ref{convergence}a shows training perplexity as a function of iterations for PGS, PBP and CE-PBP on KOS.
In the first iterations,
CE-PBP converges slower than PBP,
with the largest perplexity gap near $400$.
The gap quickly decreases with more iterations so that the training perplexity overlaps after $100$ iterations.
The perplexity gap remains within $10$ after $200$ iterations for the accuracy drop within $1\%$,
acceptable in most applications.
Fig.~\ref{convergence}b shows the training perplexity as a function of training time for PGS, PBP and CE-PBP on KOS.
CE-PBP is almost twice faster than PBP and $20\%$ faster than PGS.
In addition,
CE-PBP achieves almost the same training perplexity as PBP,
which is much lower than that of PGS.

\begin{figure}[htb]
  \centering
  \subfloat{\includegraphics[scale=0.63]{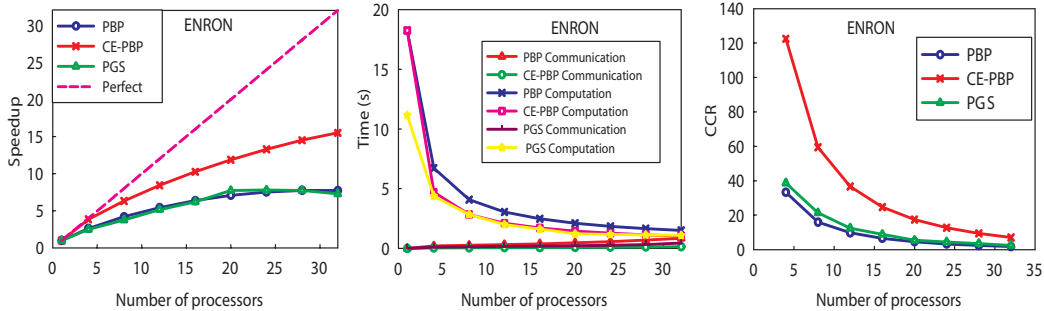}}\qquad
  \caption{(a) Left: Speedup performance.
  (b) Middle: Computation time and communication time.
  (c) Right: Parallel efficiency measured by CCR.}
  \label{ss}
\end{figure}

Fig.~\ref{ss}a illustrates the speedup performance of PGS, PBP and CE-PBP on ENRON.
The speedup is measured by $T_0/(T_0/M + T_c)$,
where $M$ is the number of processors,
$T_0$ denotes the training time of GS or BP on a single processor,
$T_c$ denotes communication cost.
Fig.~\ref{ss}a shows that CE-PBP exhibits much better speedup than PBP and PGS.
Fig.~\ref{ss}b shows the corresponding computation time and communication time.
CE-PBP and PGS have almost the same computation time,
but the former uses significantly smaller communication time than the latter.
Fig.~\ref{ss}c shows the computation/communication ratio (CCR) for the parallel efficiency of CE-PBP.
The CCR of CE-PBP is as $2$ to $3$ times as that of PBP and PGS,
reflecting a much better parallel topic modeling efficiency.

Recently,
Google reports an improved version of PGS called PGS+\cite{liu2011plda+}.
which reduces communication cost using four interdependent strategies including data placement,
pipeline processing,
word bundling and priority-based scheduling.
The ratio of communication time of both PGS+ and CE-PBP to their original algorithms, PGS and PBP, should be a fair comparison.
While a communication reduction ratio of $27.5\%$ is reported by PGS+ ($3.68$ seconds and $13.38$ seconds for PGS+ and PGS with the same settings),
we achieve a much lower ratio of about $15\%$.
Besides,
CE-PBP has a lower predictive perplexity than PGS/PGS+,

\section{Conclusions}

To reduce the communication cost that severely affects scalability in parallel topic modeling,
we have proposed CE-PBP that combines the parallel belief propagation (PBP) and a Zipf's law solution for different communication rates.
Extensive experiments on different data sets confirm that CE-PBP
is faster,
more accurate and efficient than the state-of-the-art PGS algorithm.
Since many types of data studied in the physical and social sciences can be approximated by Zipf's law,
our approach may provide a new way to accelerate other parallel algorithms.
In future work,
we shall study how to reduce the size $K$ of the global parameter matrix $\hat{\boldsymbol{\phi}}_{W \times K}$ in communication.
Also,
we plan to extend CE-PBP algorithm to learn more complicated topic models such as hierarchical Dirichlet process (HDP)~\cite{Newman:09}.

\bibliographystyle{unsrt}
\bibliography{nips2012,RBP}

\end{document}